\documentclass[10pt,twocolumn,letterpaper]{article}

\usepackage{cvpr} % uncomment this for the final submission
\usepackage{times}
\usepackage{epsfig}
\usepackage{graphicx}
\usepackage{amsmath}
\usepackage{amssymb}
\usepackage{textcomp}
\usepackage{longtable}
\usepackage{multirow}
\usepackage{booktabs}

\begin{document}

%%%%%%%%% TITLE
\title{TEMPEST: Temporal Embeddings for Scalable Driver Identification via Angular Margin Learning}

\author{Kyle Musgrove\\
The University of Tennessee, Knoxville\\
{\tt\small kmusgro1@utk.edu}
% For a paper whose authors are all at the same institution,
% omit the following lines up until the closing ``}''.
% Additional authors and addresses can be added with ``\and'',
% just like the second author.
% To save space, use either the email address or home page, not both
\and
Dylan B. Lewis\\
The University of Tennessee, Knoxville\\
{\tt\small dlewis37@utk.edu}
\and
Sarah Powers\\
Oak Ridge National Laboratory\\
{\tt\small powersss@ornl.gov}
\and
Emma J. Reid\\
Oak Ridge National Laboratory\\
{\tt\small reidej@ornl.gov}
\and
Hector Santos-Villalobos\\
The University of Tennessee, Knoxville\\
{\tt\small hsantosv@utk.edu}
}

\maketitle
\thispagestyle{empty}

%%%%%%%%% ABSTRACT
\begin{abstract}
Scalable driver identification requires embedding models that maintain discriminative performance as fleet size grows, yet existing triplet-loss formulations degrade rapidly with driver pool size and overfit to session-specific patterns under rigorous temporal evaluation. We introduce TEMPEST, a Temporal Convolutional Network embedding model trained with an additive angular margin (ArcFace) loss that enforces global class-level separation in a normalized angular space. TEMPEST maps 60-second multimodal driving windows to compact 96-dimensional embeddings, supporting truly dynamic enrollment without any retraining or classifier refitting. Under rigorous temporal evaluation on a 45-driver dataset, TEMPEST achieves 91.71\% Rank-1 accuracy, outperforming the best classical model by 17.9~pp and the strongest triplet-loss baseline by 58.4~pp. TEMPEST degrades by only 4.3~pp when growing the subject pool from 10 to 45 drivers, compared to 22~pp and 32.5~pp for supervised and unsupervised triplet-loss baselines, and its cross-session advantage is corroborated on the public KIA Soul dataset, where it outperforms the best classical model by 7.3~pp within-session and 14.3~pp cross-session. With 720K parameters, a 2.80~MB footprint, and 50-epoch convergence, TEMPEST establishes a rigorous, reproducible baseline for scalable behavioral driver biometric identification.
\end{abstract}

\begin{figure*}[t]
    \centering
    \includegraphics[width=0.95\linewidth]{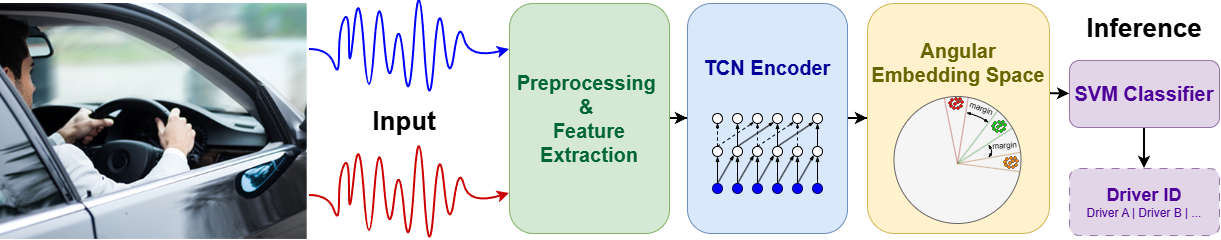}
    \caption{Driver identification pipeline. Raw multimodal data streams, behavioral signals from the vehicle CAN bus (steering, braking, acceleration), and physiological signals (heart rate) are preprocessed through filtering and windowing, transformed into feature vectors, and classified by the identification model to produce a driver identity decision.}
    \label{fig:pipeline_overview}
\end{figure*}

%%%%%%%%% BODY TEXT
\section{Introduction}

Safely transporting materials and operating secure vehicle fleets requires continuous driver identification during operation. Beyond traditional biometrics, driving behavior itself encodes person-specific patterns that persist across time, forming a behavioral biometric observable through vehicle and physiological sensors. Prior studies have shown that throttle usage, braking dynamics, steering patterns, and physiological responses can jointly characterize individual drivers \cite{Enev2016, Miyajima2007,Choi07}.

Embedding-based methods have emerged as a promising direction for driver identification because they decouple the learned representation from the number of enrolled drivers, enabling dynamic fleet enrollment without full model retraining. However, existing embedding approaches typically rely on triplet-loss or contrastive objectives that impose only local pairwise constraints \cite{DriverRep2020,driver2vec}. In practice, these formulations suffer from training instability, sensitivity to sample mining, and degraded performance as the driver pool grows \cite{wevj13110207},\cite{Khan2022}. Moreover, many reported results in the literature rely on evaluation protocols that leak temporal or session-specific information, inflating accuracy estimates and obscuring genuine model improvements \cite{Zeng22}. 

%Concurrent work by Patnala et al. \cite{TruckIDMe2025} on the DriverID dataset reported near-perfect classification accuracy (99.3\%) using features such as aftertreatment temperatures and diesel exhaust fluid tank volume signals that encode vehicle state rather than driving behavior, further illustrating the need for careful feature curation.

This paper introduces Temporal Embeddings for Multi-modal Pattern Extraction on Sequential Tasks (TEMPEST), a Temporal Convolutional Network-inspired embedding model that replaces triplet-loss training with an additive angular margin (ArcFace) objective adapted from face recognition \cite{ArcFace2019} to address limitations on previous art. We benchmark TEMPEST against five classical machine learning models and two triplet-loss deep learning baselines on the 45-driver DriverID dataset \cite{driverdata} and the Hacking and Countermeasure Research Lab (HCRL) KIA Soul dataset \cite{KIASoul2017}. 
%under chronological temporal evaluation. We further evaluate on the public , which provides four sessions per driver, enabling cross-session generalization analysis not possible with the single-session DriverID collection. 
Our primary contributions are: 

\begin{enumerate}
    \item We introduce TEMPEST, a behavioral biometric embedding encoding model trained with an additive angular margin (ArcFace) loss that enforces global class-level separation in a normalized angular space, yielding 91.71\% Rank-1 accuracy on a 45-driver dataset under rigorous temporal evaluation; a 17.4 percentage point (pp) improvement over the best classical baseline and 33.8 pp over the strongest triplet-loss baseline.

    \item We demonstrate that TEMPEST scales gracefully from 10 to 45 drivers with only 4.3 pp accuracy degradation, compared to 22 pp and 32.5 pp for supervised and unsupervised triplet-loss baselines, respectively, establishing angular margin training as a more scalable objective for behavioral biometric identification.

    \item We conduct a rigorous benchmarking study across two independent datasets, quantifying the effect of temporal data leakage and showing that random cross-validation inflates accuracy by up to 80 pp relative to chronological and cross-session evaluation, establishing reproducible baselines for the community.

    \item We show that TEMPEST's learned embeddings are sufficiently 
    discriminative for open-enrollment identification via cosine similarity 
    matching alone, achieving 91.71\% Rank-1 accuracy without any 
    pool-dependent classifier. This training-free inference scheme supports 
    truly dynamic enrollment where new drivers can be added or removed by 
    simply updating the embedding gallery, with no retraining or refitting 
    of any kind.
\end{enumerate}

% (1) we propose TEMPEST, a Temporal Convolutional Network (TCN) based embedding model trained with ArcFace loss for driver identification; (2) we provide a systematic comparison of unsupervised triplet loss, supervised triplet loss, and angular margin objectives alongside classical baselines; (3) we demonstrate that TEMPEST scales gracefully from 10 to 45 drivers with minimal accuracy degradation; and (4) we show that TEMPEST's decoupled encoder-SVM architecture enables practical deployment with dynamic fleet enrollment.

%%%%%%%%%%%%%%%%%%%%%%%%%%%%%%%%%%%%%%%%%%%%%%%%%%%%%%%%%%%%%%%

\section{Related Work}
Driving style has long been studied as a behavioral biometric. Early work demonstrated the feasibility of using classical models such as Hidden Markov Models, Gaussian Mixture Models, and decision trees on small driver cohorts and fixed routes \cite{Miyajima2007,Wakita06,Wahab09}. Subsequent studies leveraged Controller Area Network (CAN) bus and On-Board Diagnostic II (OBD-II) signals and introduced public datasets such as the HCRL KIA Soul dataset \cite{KIASoul2017}, enabling systematic comparisons across methods \cite{Ezzini18,Mekki,Xun_2020_TII}. However, early classical methods assume a fixed subject pool and require retraining the model to accommodate new drivers.

As datasets have grown, deep learning approaches have emerged, including convolutional and recurrent models operating on multivariate time series \cite{Zhang_2019_Sensors,Zhengping,girma2019driveridentificationbasedvehicle,ABDENNOUR2021103707,Alqahtani}. More recently, embedding-based methods have been proposed that learn compact driver representations using triplet loss or contrastive objectives, often implemented with causal convolutions \cite{Hallac18,driver2vec,Hallac16}. Such embeddings support a dynamic subject pool in which gallery identities can be added or removed on demand. While these methods show promise, reported accuracies often degrade as the number of drivers increases or when models are evaluated across sessions \cite{wevj13110207},\cite{Khan2022},\cite{khan2022machinelearningapproachdriver}.

Embedding-based approaches such as Driver2Vec \cite{driver2vec} and Drive2Vec \cite{Hallac16} attempt to address scalability through learned representations, but their triplet-loss formulations provide only local pairwise constraints and do not enforce global class-level structure. In facial recognition, angular margin losses such as ArcFace \cite{ArcFace2019} have been shown to produce highly discriminative embeddings by enforcing a fixed angular margin between class centers on a hypersphere. To our knowledge, TEMPEST is the first application of angular margin learning to behavioral biometric identification from multimodal driving signals.

%%%%%%%%%%%%%%%%%%%%%%%%%%%%%%%%%%%%%%%%%%%%%%%%%%%%%%%%%%%%%%%

\section{Datasets}
We evaluate our framework on two different datasets that differ in size, number of sessions, vehicle type, and sensor configuration. The first is a newly curated dataset of 45 real-world drivers derived from an Oak Ridge National Laboratory collection \cite{driverdata}, which we cleaned, processed, and refer to as the DriverID dataset. The second is the widely used HCRL KIA Soul dataset \cite{kia2016}, which serves as a public benchmark for comparison with prior work.

\subsection{ORNL Driver Identification Dataset}
The Oak Ridge National Laboratory Driver Identification (DriverID) dataset is derived from a real-world collection of 50 volunteer truck drivers \cite{driverdata}. Details on the instrumentation and data collection process are documented by Biggs et al. \cite{Biggs2024}. The primary purpose of this study is to observe participants’ behavior during and after a cybersecurity attack on the vehicle. The drivers were divided into three experimental groups based on the level of advance warning they received about the attack. Group 1 (G1, 17 Subjects) received no warning of the attack. Group 2 (G2, 16 subjects) was informed that a cyberattack may or may not occur during the drive. Group 3 (G3, 17 Subjects) received both a warning that the attack would occur and explicit instructions to slow down and pull over to the side of the road if it did. 

Each driver completed a continuous 40–60 minute drive in the same instrumented vehicle on a fixed route. The vehicle was instrumented with its native CAN bus \cite{canbus} (e.g., speedometer, throttle position, brake pressure) and an aftermarket Racelogic VBOX 3i \cite{vbox} for precise GPS-based and inertial motion measurements. Participants wore an Empatica E4 wristband \cite{hr} with a heart rate monitor. All data streams were timestamped and aligned to form a unified multimodal time series for each driver, with 816 raw features.

After cleaning, processing, and adaptation for driver identification, the dataset comprises 45 drivers (15 from G1, 14 from G2, and 16 from G3). Although the three groups experienced different levels of cyberattack awareness, our analysis did not reveal a systematic biometric difference between groups: the most confused driver pairs span across group boundaries rather than clustering within groups, and no group exhibited consistently higher or lower identification accuracy. After filtering, resampling to a common 1 Hz rate, and removing signals encoding vehicle state rather than driving behavior (e.g., GPS coordinates, aftertreatment temperatures), 19 behavioral features were retained.

% A detailed treatment of the preprocessing pipeline and complete feature enumeration will be presented in a forthcoming companion paper focused on dataset curation and signal quality analysis.

\subsection{HCRL KIA Soul Dataset}
The HCRL Kia Soul dataset \cite{kia2016} serves as our benchmark for corroborating and comparing model performance with prior work. Before analysis, we reorganized the data by driver identity and by temporally ordering the samples. Each of the 10 drivers completed four distinct driving sessions, and we assigned explicit session labels (Drive 1–4) to each recording, ensuring that these labels were consistently attached to the corresponding data. These session annotations enable clean data splits for the training and testing sets, preventing samples from the same driving session from appearing in both sets. The dataset contains 54 features, but only 15 are considered important by previous work \cite{kia2016}. This multi-session structure of the dataset is critical for our evaluation: it allows for Leave-One-Drive-Out (LODO) protocols that hold out an entire session, directly testing whether models generalize across multiple sessions rather than relying on a single within-session patterns.

\section{Modeling Approach}

Classical classifiers operate on hand-engineered feature representations and learn decision boundaries directly in the original statistical feature space. Although effective under controlled conditions, this formulation tightly couples model performance to the stability of the features and makes these methods vulnerable to temporal leakage and session-specific artifacts. When evaluation protocols enforce chronological or cross-session separation, classical classifiers tend to rely more heavily on spurious contextual cues, which in turn degrades generalization. Moreover, these approaches implicitly assume a fixed, closed set of enrolled drivers, thereby limiting scalability.

Embedding-based methods mitigate these limitations by learning driver representations rather than performing explicit multi-class classification. Instead of mapping window-level summaries to discrete class labels, these methods learn an embedding space, in which embeddings from the same driver are encouraged to occupy compact regions, while embeddings from different drivers are separated by margin-based constraints. This formulation decouples identity modeling from the number of enrolled drivers and naturally supports scalable evaluation under cross-session conditions.

\subsection{DriverRep Baseline}
We adopt DriverRep \cite{DriverRep2020} as our embedding-based deep learning baseline. DriverRep uses a Temporal Convolutional Network (TCN) \cite{TCN2018},\cite{wavenet} with dilated causal convolutions and residual connections to map fixed-length driving windows to $\mathbb{R}^{128}$ driver embeddings. The model is trained with a triplet loss \cite{FaceNet} that encourages embeddings from the same driver to be more similar to each other than to embeddings from different drivers. At inference time, identity decisions are obtained via similarity-based matching or by applying downstream classifiers in the embedding space \cite{MARTINELLI2020102504}. This representation-centric formulation offers a principled alternative to feature-driven classical classifiers and facilitates a direct analysis of cross-session robustness.

\subsection{TEMPEST}

Although DriverRep effectively learns relative similarities, its triplet-loss formulation imposes local constraints that depend on the selection of informative triplets. In practice, this dependence introduces training instability and complicates scaling as the number of drivers increases, since the embedding geometry is shaped primarily by pairwise or triplet-level relationships. Moreover, triplet loss does not explicitly enforce compact class-level structure, which can result in overlapping driver embeddings when intra-driver variability is high, as is common in real-world driving data collected across time, routes, and conditions.

To address these limitations, we introduce the DriverRep-inspired technique TEMPEST. The technique replaces the triplet-loss objective with an additive angular margin loss, specifically ArcFace \cite{ArcFace2019}, reframing driver identification as a supervised discriminative embedding problem. By operating in a normalized angular space and enforcing a fixed margin at the class level, ArcFace introduces global constraints that promote consistent separation between driver embeddings across sessions. Unlike triplet-based objectives, angular margin–based supervision eliminates the need for explicit sample mining and yields stable convergence under standard classification training pipelines. These properties make TEMPEST particularly well-suited for behavioral biometric settings, where subtle temporal patterns must remain discriminative under cross-session and longitudinal evaluation.

In the original DriverRep implementation, the authors adopt an unsupervised triplet loss approach in which positive samples are shorter subsequences of the anchor sample, while negative samples are drawn from other drivers’ sessions. In our application, the drivers are known; we therefore opt for a fully supervised approach instead, where the anchor, positive, and negative samples are given by $\mathbf{X}_{d,r}^{(i)}$, $\mathbf{X}_{d,r}^{(j)}$, $\mathbf{X}_{\tilde{d},r}^{(k)}$, respectively, where $i \neq j$ and $d \neq \tilde{d}$. 

\subsubsection{Encoder Architecture} 
\begin{figure}[h]
    \centering
    \includegraphics[width=0.99\linewidth]{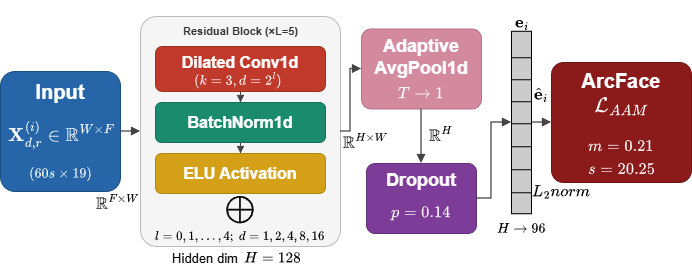}
    \caption{TEMPEST architecture, $f_\theta(\cdot)$ with the following components: (1) Multivariate driving windows $X^{(i)}_{d,r} \in \mathbb{R}^{W \times F}$ as input, (2) a TCN encoder with a single ($B=1$) block of five ($L=5$) exponentially dilated layers, (3) an adaptive average pooling, (4) a final linear projection to produce an $L_2$-normalized ($\hat{e}_i$) embedding $\in \mathbb{R}^{96}$, and (5) the ArcFace objective $\mathcal{L}_{AAM}$ for training.}
    \label{fig:tempest_arch}
\end{figure}

TEMPEST employs a TCN encoder, illustrated in Figure~\ref{fig:tempest_arch}, consisting of stacked residual blocks with exponentially dilated causal convolutions. Each block contains three sequential operations: a dilated 1D convolution with a kernel size $k = 3$, batch normalization, and an ELU activation, followed by dropout. The dilation factor increases exponentially across layers as $2^l$ for layer $l \in \{0, 1, \ldots, L-1\}$, where $L = 5$, providing a receptive field that spans the full 60-second input window. A residual connection with a $1\times1$ projection bypasses each block to facilitate gradient flow. Given an input window $\mathbf{X}^{(i)}_{d,r} \in \mathbb{R}^{W \times F}$, the encoder first transposes the input to channel-first format $\mathbb{R}^{F \times W}$, processes it through $B = 1$ block of $L = 5$ dilated layers, applies adaptive average pooling over the temporal axis, and projects the result through a fully connected layer to produce an embedding $\mathbf{e} \in \mathbb{R}^{96}$. TEMPEST uses a 96-dimensional embedding in place of DriverRep's  128-dimensional representation, as preliminary experiments showed only marginal accuracy gains from the larger embedding at the cost of increased model size.

\subsubsection{ArcFace Loss Formulation} TEMPEST replaces the triplet-loss objective with the Additive Angular Margin (ArcFace) loss. Let $\mathbf{e}_i = f_\theta(\mathbf{X}^{(i)}_{d,r})$ denote the $L_2$-normalized embedding for window $i$, and let $\mathbf{W} = [\mathbf{w}_1, \ldots, \mathbf{w}_C]$ be the $L_2$-normalized class weight matrix, where $C$ is the number of enrolled drivers. The cosine similarity between embedding $\mathbf{e}_i$ and class center $\mathbf{w}_j$ is $\cos\theta_j = \mathbf{e}_i^\top \mathbf{w}_j$. The ArcFace loss adds an angular margin $m$ to the target angle:

\begin{equation}
    \mathcal{L}_{\text{AAM}} = -\log \frac{e^{s \cos(\theta_{y_i} + m)}}{e^{s \cos(\theta_{y_i} + m)} + \sum_{j \neq y_i} e^{s \cos\theta_j}}
    \label{eq:arcface}
\end{equation}

\noindent where $\theta_{y_i}$ is the angle between embedding $\mathbf{e}_i$ and its true class center $\mathbf{w}_{y_i}$, where $m=0.21$ rad and $s=20.25$ are hyperparameters selected via Optuna Bayesian optimization on a held-out validation split drawn exclusively from the training portion (no test data was accessed during tuning). The margin $m$ enforces a minimum angular separation between driver embeddings during training, producing tighter intra-class clusters and wider inter-class boundaries compared to standard softmax classification.

At inference time, we evaluate the TEMPEST L2-normalized embeddings under two configurations. 
In the first, which we refer to as \textbf{TEMPEST-Raw}, driver identity is determined by nearest neighbor retrieval using cosine distance against a gallery of enrolled driver embeddings, requiring no trainable parameters beyond the encoder itself. 
In the second, which we refer to as \textbf{TEMPEST-SVM}, a lightweight SVM is trained on the precomputed embeddings of the enrolled drivers and used as the classification head at inference. An RBF-kernel SVM with $C=1.0$
(i.e., default sklearn settings) was used for all models.
Both configurations share the same frozen encoder; the distinction lies solely in how the embedding is matched to an identity. 
All three deep learning models (TEMPEST, Sup-TL, and DriverRep) are evaluated under the SVM configuration.

\section{Results and Discussion}
All deep learning experiments on the DriverID dataset were run 10 times with different random seeds. MLP was also evaluated across 10 independent runs. We report the best accuracy for the KIA dataset and the mean accuracy and standard deviation ($\sigma$) across the DriverID dataset unless otherwise noted. Classical models (DT, RF, kNN, XGB) were evaluated in a single run given their negligible variance, and serve primarily to illustrate how low-capacity models can appear competitive under leaky evaluation protocols.

DriverRep is evaluated under two training configurations: the original \textbf{DriverRep} formulation using the unsupervised triplet loss described in~\cite{DriverRep2020}, and a supervised variant (\textbf{Sup-TL}) in which anchor, positive, and negative samples are drawn using ground-truth driver labels, matching the supervision strategy used in TEMPEST. Together with TEMPEST, these three configurations form a controlled ablation over the training objective on a fixed TCN encoder.

We define two leakage-aware evaluation protocols. The first is a within-session chronological split, using the first 80\% of each drive for training and the remaining 20\% for testing. The second is a cross-session leave-one-drive-out (LODO) protocol, where entire sessions are held out for evaluation. For comparison, we also report results using random cross-validation to illustrate data leakage effects.

\subsection{Evaluation Protocol Validation and Baseline Reproduction}

To validate our deep learning baseline, we re-implemented the DriverRep model \cite{DriverRep2020} following the original authors' methodology on the 10-driver KIA dataset and 128-dimensional embedding. Under their evaluation protocol, our implementation achieved an average classification accuracy of 81.50\%, closely matching the 81.70\% reported in the original work. However, after applying our strict temporal evaluation framework (i.e., Session Split) and removing session-specific (leaky) features, DriverRep's performance dropped to approximately 25\% on average for the KIA 10-driver task.

Table~\ref{tab:evaluation_impact} reports model performance under different evaluation methodologies for both datasets, using a 96-dimensional embedding for all deep learning models. The results confirm the effect of temporal leakage: random 10-fold cross-validation yields near-perfect accuracy ($\approx$99\%) across all models regardless of architecture, whereas chronological and cross-session evaluation yield substantially lower, more realistic estimates, exposing differences in generalization that random splitting conceals. 
 On the DriverID dataset, DriverRep collapses from 99.89\% to 33.35\% under chronological splitting, a drop of 66.5 pp, whereas TEMPEST retains 91.21\%, a 57.9 pp gap between the two models that random splitting completely conceals. The effect is even more pronounced under cross-session evaluation on the KIA dataset, where DriverRep drops to 17.36\% approaching chance level, while TEMPEST retains 40.93\%.

\begin{table}[h]
    \centering
    \caption{Combined evaluation methodology impact across datasets. Results are from the models' best performance.}
    \label{tab:evaluation_impact}
    \resizebox{0.8\columnwidth}{!}{%
    \begin{tabular}{l l l l c}
        \toprule
        \textbf{Dataset} & \textbf{MLP} & \textbf{DriverRep} & \textbf{Sup-TL} & \textbf{TEMPEST}\\
        \midrule
        \multicolumn{5}{l}{\textbf{KIA Dataset}} \\
        Random Split & 99.92 & 99.51 & 99.89 & \textbf{99.99} \\
        \textit{Chronological Split} \\
        \ Drive 1 & 86.22 & 89.50 & 90.86 & \textbf{93.60} \\
        \ Drive 2 & 69.62 & 67.21 & 74.96 & \textbf{78.67} \\
        \ Drive 3 & 80.67 & 78.41 & 85.12 & \textbf{87.05} \\
        \ Drive 4 & 81.15 & 78.30 & 79.58 & \textbf{87.39} \\
        \textit{Session Split} \\
        \ D2,3,4$\rightarrow$D1 & 44.65 & 26.54 & 46.89 & \textbf{63.97} \\
        \ D1,3,4$\rightarrow$D2 & 25.02 & 17.36 & 31.19 & \textbf{40.93} \\
        \ D1,2,4$\rightarrow$D3 & 54.90 & 20.48 & 35.60 & \textbf{65.96} \\
        \ D1,2,3$\rightarrow$D4 & 51.63 & 36.89 & 39.69 & \textbf{62.34} \\
        \midrule
        \multicolumn{5}{l}{\textbf{DriverID Dataset}} \\
        Random Split & 99.99 & 99.89 & 99.98 & \textbf{99.99}\\
        Chronological Split & 73.86 & 33.35 & 57.39 & \textbf{91.21}\\
        \bottomrule
    \end{tabular}
    }
\end{table}

\subsubsection{Efficiency–Accuracy Trade-offs}
\begin{figure}[h]
    \centering
    \includegraphics[width=0.75\columnwidth]{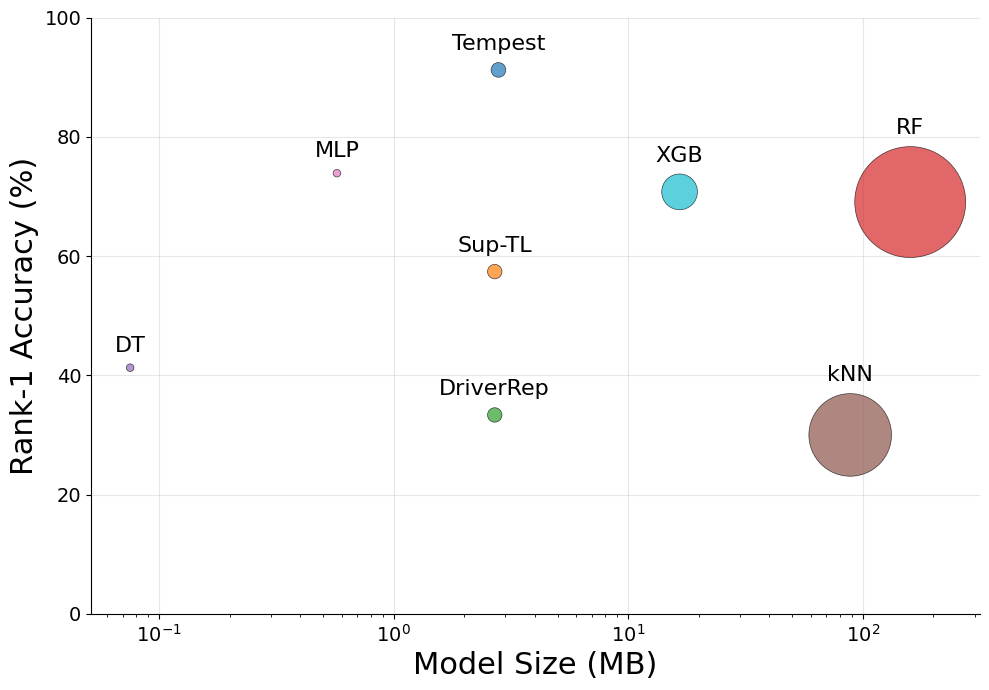}
    \caption{Model size vs.\ Rank-1 accuracy for all evaluated models on the 45-driver DriverID dataset (  chronological split ). Marker size reflects model footprint.}
    \label{fig:efficiency}
\end{figure}

Figure~\ref{fig:efficiency} plots Rank-1 accuracy against memory footprint for all evaluated models on the 45-driver DriverID dataset. TEMPEST achieves the highest accuracy (91.71\% Raw, 91.21\% SVM) at only 2.80~MB, while the classical models that come closest in accuracy, MLP (73.89\%) and XGB (70.76\%), require 574~KB and 16.6~MB, respectively, with Random Forest and kNN demanding 159~MB and 88~MB for far lower accuracy. Critically, DriverRep and Sup-TL share nearly the same footprint as TEMPEST ($\approx$2.7~MB) yet reach only 33.35\% and 57.39\% respectively, isolating the ArcFace loss rather than model capacity as the source of TEMPEST's advantage. The comparable Raw and SVM accuracies (91.71\% vs. 91.21\%) further confirm that discriminative power resides in the embeddings themselves, independent of the downstream classifier.

\subsubsection{Supervised Triplet Loss vs. Unsupervised Triplet Loss}
\begin{figure}[h]
    \centering
    \includegraphics[width=0.8\columnwidth]{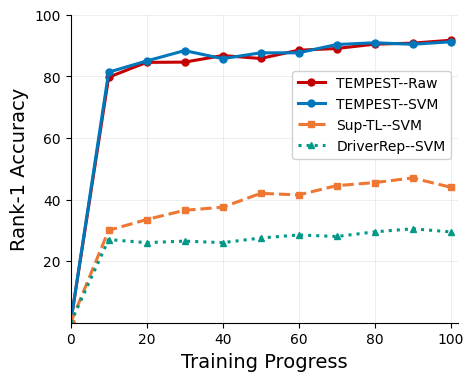}
    \caption{Rank-1 SVM test accuracy as a function of training progress for the three deep learning models on the 45-driver DriverID dataset.}
    \label{fig:contrastive_supervised}
\end{figure}

Figure~\ref{fig:contrastive_supervised} compares the convergence behavior of TEMPEST, Sup-TL, and DriverRep, which share the same TCN encoder but differ only in training objective. TEMPEST reaches 81\% within 5 epochs and converges to 91\% by epoch 50, while Sup-TL and DriverRep plateau at 57\% and 33\% respectively after 400 epochs. Replacing unsupervised with supervised triplet loss (DriverRep $\rightarrow$ Sup-TL) yields a modest 24 pp improvement, whereas replacing triplet loss with ArcFace (Sup-TL $\rightarrow$ TEMPEST) yields an additional 34 pp gain; confirming that the choice of loss function, not the supervision strategy, is the dominant factor. This advantage stems from ArcFace's softmax-based formulation, which produces gradients for all $C = 45$ driver classes simultaneously at every step, whereas triplet-based losses update only the anchor and negative driver identities per triplet, leaving the remaining 43 drivers without a gradient signal. The result is denser supervision, faster convergence, and a substantially higher accuracy ceiling.

\subsection{Deep Learning Model Analysis}
We examined the three deep learning models TEMPEST, Sup-TL, and DriverRep along four dimensions: scalability with respect to driver pool size, rank-based biometric evaluation, per-driver identification quality, and embedding space structure. All experiments use the 45-driver DriverID dataset under the chronological split, with 10 independent trials per configuration to quantify variance. For each trial, driver subsets are sampled randomly, and accuracy is measured on the temporally held-out 20\% of each selected driver's session. 

\subsubsection{Pool Size Scaling}
\begin{figure}[h]
    \centering
    \includegraphics[width=0.8\columnwidth]{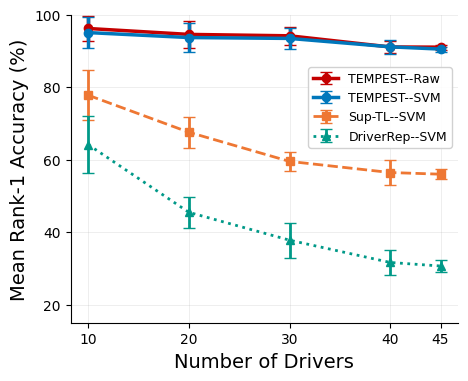}
    \caption{Mean Rank-1 SVM test accuracy for driver pool size. Bars indicate $\pm 1\sigma$ over 10 trials. TEMPEST maintains robust performance as the pool scales from 10 to 45 drivers.}
    \label{fig:nway}
\end{figure}

Table \ref{tab:nway} and Figure \ref{fig:nway} present the scaling behavior as the driver pool increases from 10 to 45 drivers. As expected, recognition difficulty increases with pool size, so all models degrade to some degree, while TEMPEST's advantage is in the rate of degradation. TEMPEST maintains accuracy above 90\% at all pool sizes, degrading by only 4.3 pp ($\sigma=0.70$ at 45 drivers), while DriverRep degrades by 32.5 pp and Sup-TL by 22 pp. 

%Error analysis shows that TEMPEST's residual misidentifications are concentrated among a small number of driver pairs with similar behavioral profiles, suggesting that adaptive per-pair margins or hierarchical clustering in the embedding space could further improve discrimination for these difficult cases.

% Our scalability analysis reveals that TEMPEST degrades gracefully from 10 to 45 drivers, losing only 4 percentage points in mean accuracy compared to 32.5 points for DriverRep and 21.8 points for Sup-TL. The low variance of TEMPEST at larger pool sizes ($\sigma = 0.70$ at 45 drivers) further indicates robust training dynamics. Error analysis shows that TEMPEST's residual misidentifications are concentrated among a small number of driver pairs with similar behavioral profiles, suggesting that adaptive per-pair margins or hierarchical clustering in the embedding space could further improve discrimination for these difficult cases.

\begin{table}[h]
    \centering
    \caption{DriverID N-way classification accuracy (mean $\pm$ $\sigma$ over 10 trials) for deep learning models under chronological split evaluation.}
    \label{tab:nway}
    \resizebox{0.8\columnwidth}{!}{%
    \begin{tabular}{c c c c c}
        \toprule
            & \textbf{Raw} & \multicolumn{3}{c}{\textbf{SVM}} \\
            \cmidrule(lr){2-2} \cmidrule(lr){3-5}
            \textbf{\# Drivers} & \multicolumn{2}{c}{\textbf{TEMPEST(\%)}} & \textbf{Sup-TL(\%)} & \textbf{DriverRep(\%)} \\
            \midrule
            10 & 96.20 $\pm$ 3.3 & 95.07 $\pm$ 4.4 & 77.78 $\pm$ 6.9 & 64.14 $\pm$ 7.8 \\
            20 & 94.57 $\pm$ 3.7 & 93.70 $\pm$ 4.1 & 67.56 $\pm$ 4.2 & 45.46 $\pm$ 4.3 \\
            30 & 94.18 $\pm$ 2.4 & 93.46 $\pm$ 2.8 & 59.52 $\pm$ 2.6 & 37.78 $\pm$ 4.8 \\
            40 & 91.10 $\pm$ 1.5 & 91.14 $\pm$ 1.9 & 56.45 $\pm$ 3.4 & 31.63 $\pm$ 3.4 \\
            45 & 91.05 $\pm$ 0.6 & 90.51 $\pm$ 0.7 & 55.99 $\pm$ 1.4 & 31.65 $\pm$ 1.7 \\
        \bottomrule
    \end{tabular}
    }
\end{table}

\subsubsection{Rank-Based Evaluation}
\begin{figure}[h]
    \centering
    \includegraphics[width=0.8\columnwidth]{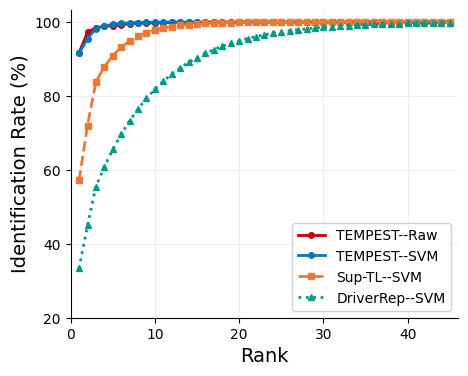}
    \caption{Cumulative Match Characteristic (CMC) curves for the three deep learning models on the 45-driver DriverID dataset under temporal evaluation.}
    \label{fig:cmc}
\end{figure}

Figure \ref{fig:cmc} presents the Cumulative Match Characteristic (CMC) curves, which measure the probability that the correct identity appears within the top-k ranked drivers. TEMPEST reaches 96\% by Rank-3 and 99\% by Rank-5. Sup-TL requires 5 ranks to reach 90\%, while DriverRep does not surpass 90\% until Rank-15, reflecting the tight angular separation enforced by ArcFace.   

\begin{figure}[h]
    \centering
    \begin{subfigure}[t]{0.45\linewidth}
        \centering
        {%
          \setlength{\fboxsep}{0pt}%
          \setlength{\fboxrule}{0.4pt}%
          \fcolorbox{black}{white}{%
            \includegraphics[width=\dimexpr\linewidth-2\fboxrule\relax]{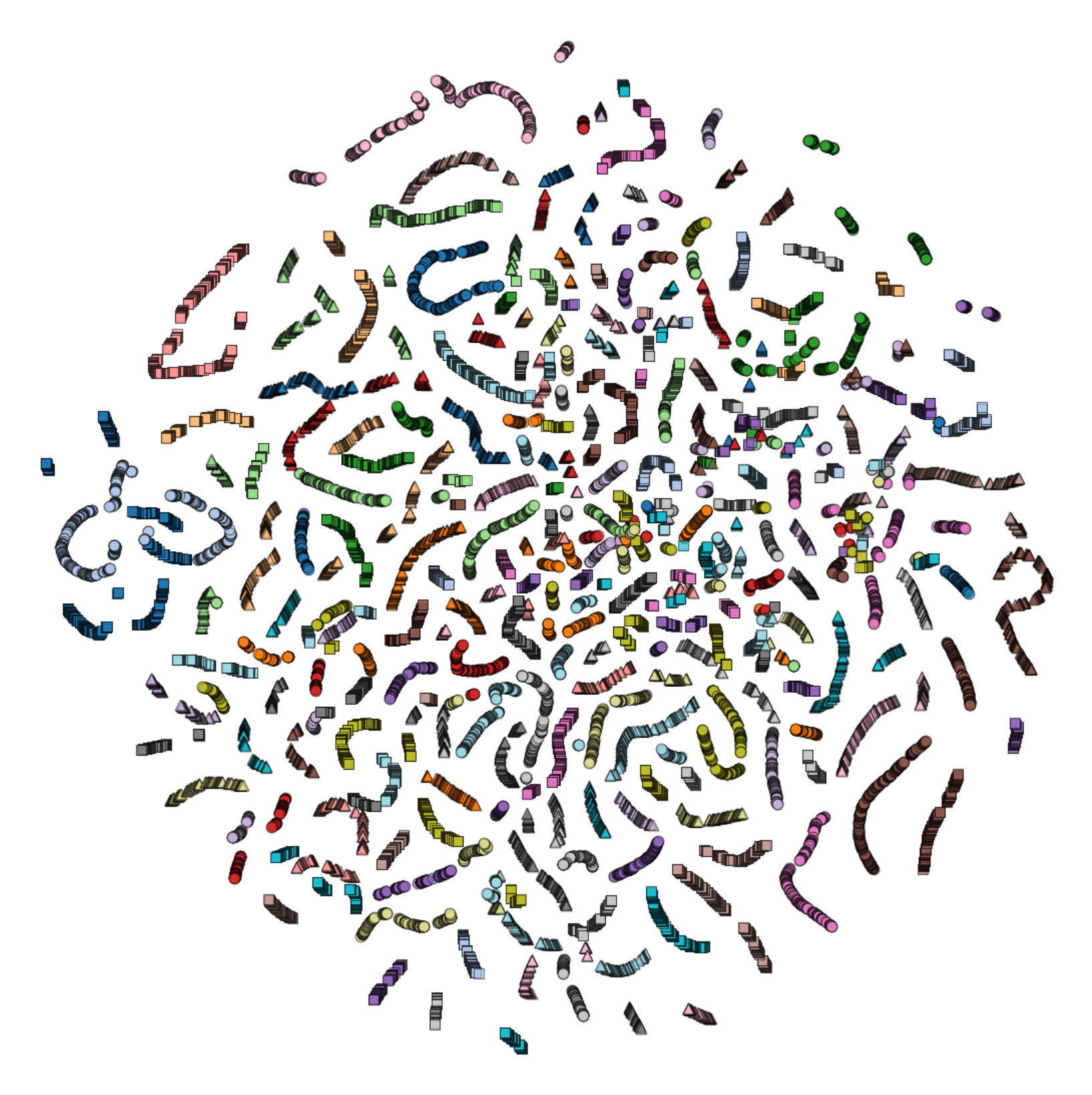}%
          }%
        }
        \caption{DriverRep embeddings}
    \end{subfigure}\hfill%
    \begin{subfigure}[t]{0.45\linewidth}
        \centering
        {%
          \setlength{\fboxsep}{0pt}%
          \setlength{\fboxrule}{0.4pt}%
          \fcolorbox{black}{white}{%
            \includegraphics[width=\dimexpr\linewidth-2\fboxrule\relax]{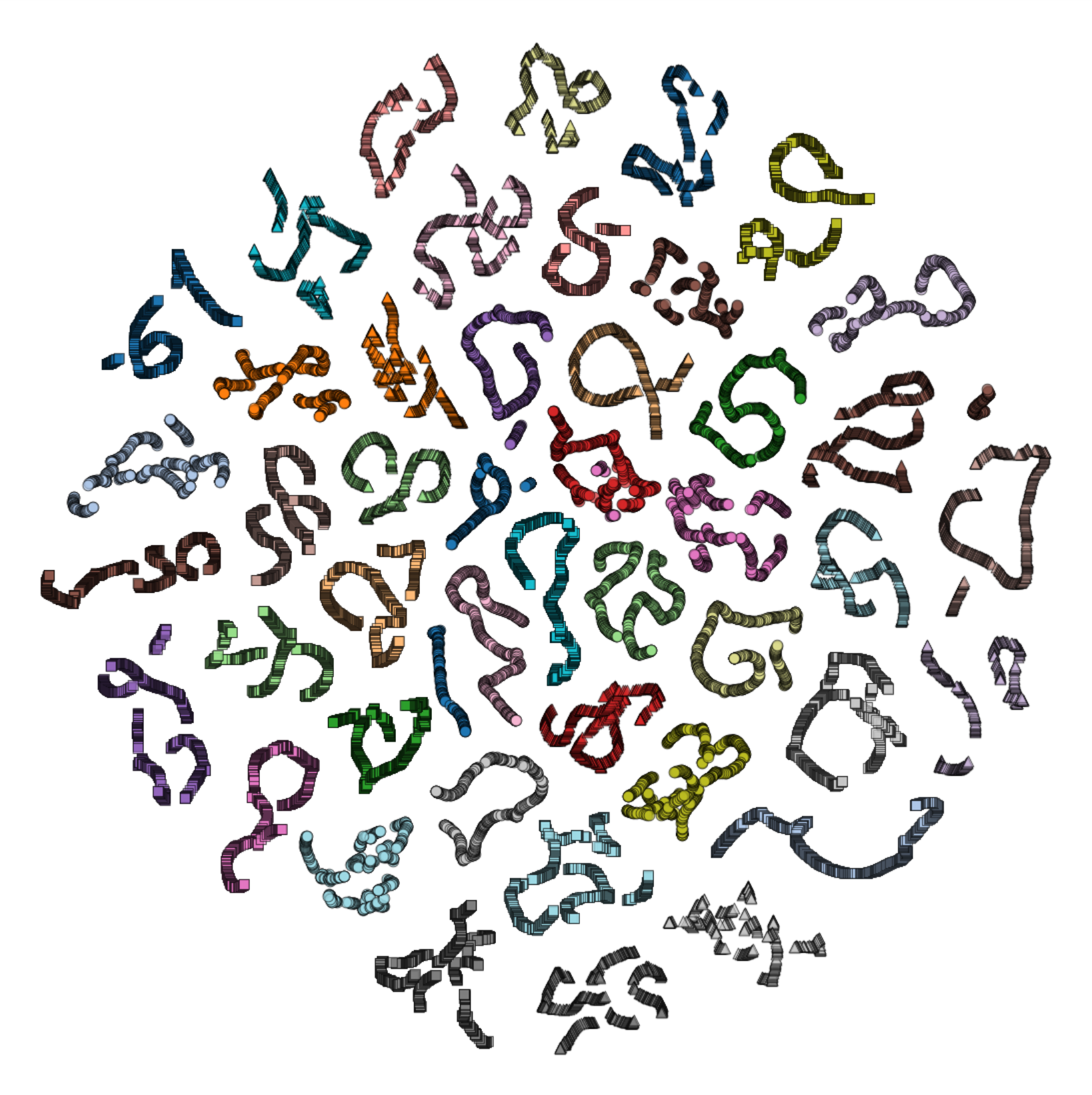}%
          }%
        }
        \caption{TEMPEST embeddings}
    \end{subfigure}
    \caption[t-SNE Visualization of Embedding Spaces]{DriverRep (left) vs. TEMPEST (right) on DriverID 45-drivers; distinct clusters under TEMPEST reflect stronger inter‑driver separation.}
    \label{fig:tsne_comparison}
\end{figure}

Figure \ref{fig:tsne_comparison} presents t-SNE \cite{tSNE2008} visualizations of the learned embedding spaces. DriverRep embeddings exhibit significant overlap among the 45 drivers, whereas TEMPEST embeddings form distinct clusters with clear inter-driver boundaries, visually confirming the angular margin advantage. This structural difference is consistent with the CMC results: the compact, well-separated clusters produced by the additive angular margin training enable high confidence identity retrieval at low ranks, whereas the DriverRep geometry requires many candidates before the correct identity surfaces.

\subsubsection{Embedding Similarity Distributions}
\begin{figure}[h]
    \centering
    \begin{subfigure}[t]{0.33\linewidth}
        \centering
        \includegraphics[width=\linewidth]{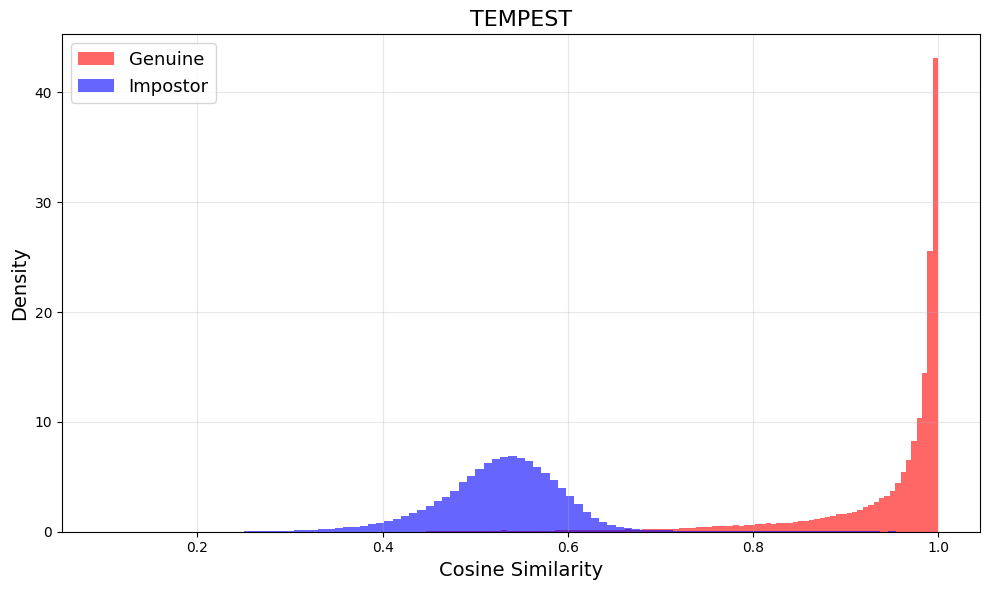}
        \caption{TEMPEST}
    \end{subfigure}\hfill%
    \begin{subfigure}[t]{0.33\linewidth}
        \centering
        \includegraphics[width=\linewidth]{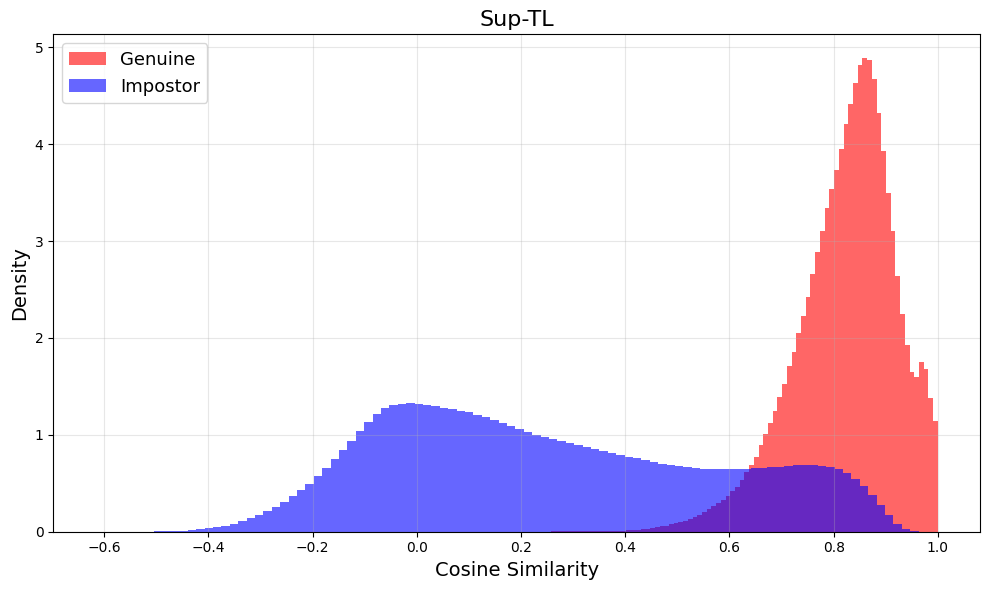}
        \caption{Sup-TL}
    \end{subfigure}\hfill%
    \begin{subfigure}[t]{0.33\linewidth}
        \centering
        \includegraphics[width=\linewidth]{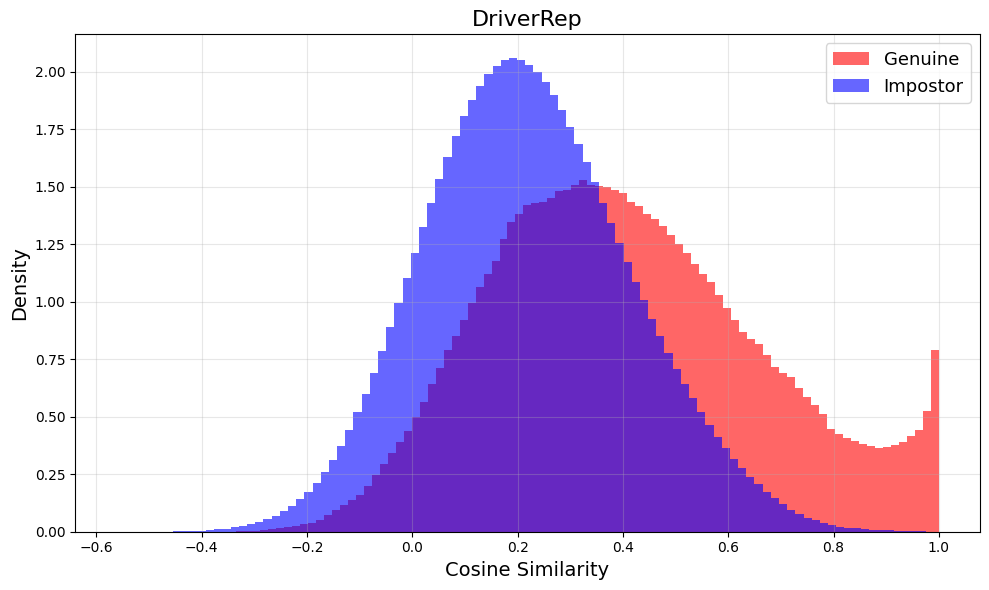}
        \caption{DriverRep}
    \end{subfigure}
    \caption{Cosine similarity distributions of genuine and impostor embedding pairs on the 45-driver DriverID dataset. TEMPEST achieves near complete separation; triplet-based models exhibit substantial overlap.}
    \label{fig:genuine_impostor}
\end{figure}

Figure \ref{fig:genuine_impostor} presents the cosine similarity distributions between genuine (same-driver) and impostor (different-driver) embedding pairs for the three deep learning models. TEMPEST produces near-complete separation: genuine pairs concentrate above 0.9 while impostor pairs peak near 0.53 with negligible overlap. Sup-TL achieves partial separation with genuine pairs peaking around 0.85, but substantial impostor mass extends into the 0.6-0.8 range. DriverRep shows extensive overlap, with both distributions centered around 0.2-0.4, explaining its poor identification performance. These distributions directly illustrate the angular margin effect: ArcFace training compresses intra-class angles toward zero while enforcing a minimum inter-class gap, yielding the bimodal separation visible in the TEMPEST panel.
 
\subsubsection{Per-Driver Identification Analysis} 
\begin{figure}[h]
    \centering
    \begin{subfigure}[t]{0.4\linewidth}
        \centering
        \includegraphics[width=\linewidth]{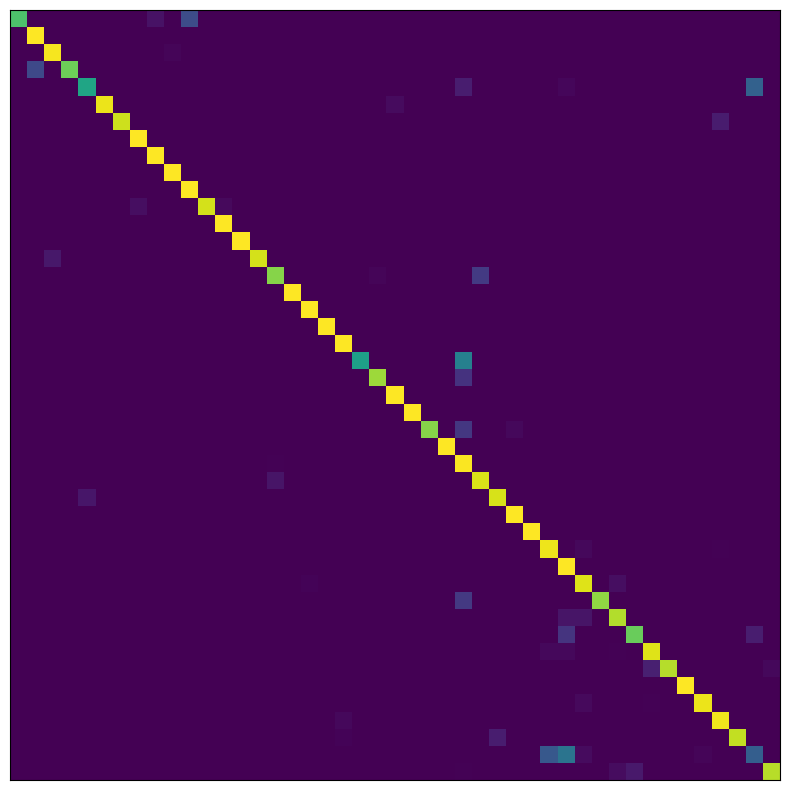}
        \caption{TEMPEST}
    \end{subfigure}\hfill%
    \begin{subfigure}[t]{0.44\linewidth}
        \centering
        \includegraphics[width=\linewidth]{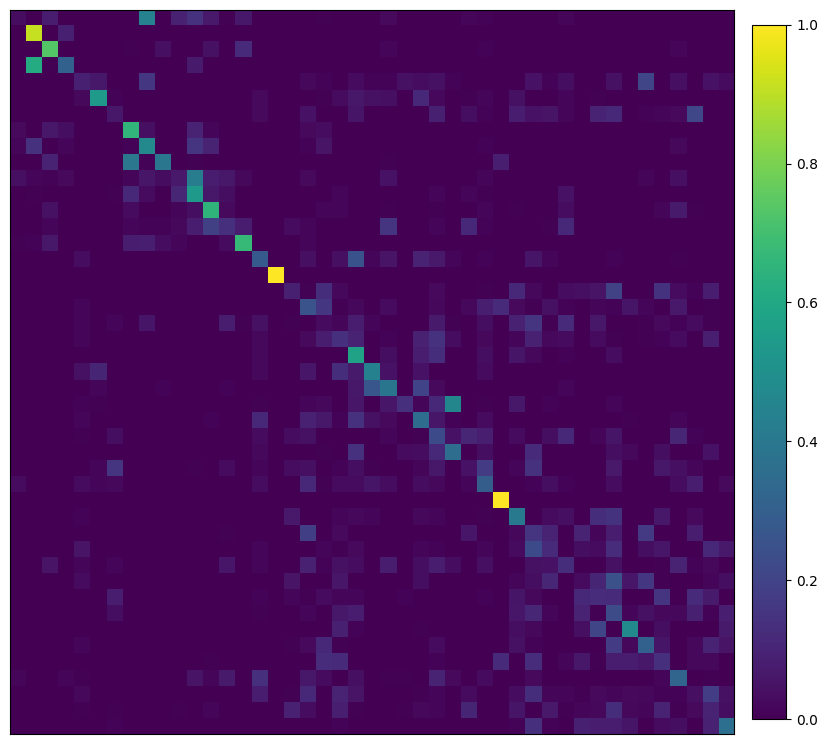}
        \caption{DriverRep}
    \end{subfigure}
    \caption{Normalized confusion matrices for the 45-driver DriverID dataset under temporal evaluation. TEMPEST produces a strong diagonal with minimal off-diagonal confusion, while DriverRep exhibits strong misclassification across the full driver pool.}
    \label{fig:cfm}
\end{figure}

Figure \ref{fig:cfm} presents the normalized confusion matrices for TEMPEST and DriverRep. TEMPEST exhibits a strong diagonal with minimal off-diagonal confusion, with most drivers achieving near perfect classification. DriverRep shows a diffuse pattern with substantial off-diagonal mass, reflecting widespread identification failure where predictions are scattered across many incorrect identities. 

Table~\ref{tab:driver_metrics} quantifies these differences across all 45 drivers. TEMPEST achieves a mean per-class F1 of 91\%, with 32 of 45 drivers individually exceeding 90\% F1, an AUC of 99.52\%, and an EER of 2.98\%, indicating robust identification performance across the full driver pool rather than a small subset of easy-to-identify drivers. In contrast, DriverRep yields a mean per-class F1 of only 31.64\%, with 35 of 45 drivers falling below 50\%, while Sup-TL occupies an intermediate position at 54.91\%, consistent with their respective positions on the ablation axis.

\begin{table}[h]
    \centering
    \caption{Driver identification metrics across deep learning models on the 45-driver DriverID dataset (best performing runs).}
    \resizebox{0.8\columnwidth}{!}{%
    \begin{tabular}{l c c c c}
    \toprule
        & \textbf{Raw} & \multicolumn{3}{c}{\textbf{SVM}} \\
        \cmidrule(lr){2-2} \cmidrule(lr){3-5}
        \textbf{Metric} & \multicolumn{2}{c}{\textbf{TEMPEST}} & \textbf{Sup-TL} & \textbf{DriverRep} \\
        \midrule
         Classification Accuracy & 91.71 & 91.21 & 57.39 & 33.35 \\
         F1 (macro) & 92.01 & 91.37 & 54.91 & 31.64 \\
         AUC (macro) & 99.30 & 99.52 & 95.25 & 87.56 \\
         EER & 1.51 & 2.98 & 11.13 & 20.57 \\
         Prec (macro) & 93.51 & 93.01 & 58.01 & 33.10 \\
         Recall (macro) & 91.82 & 91.29 & 57.18 & 32.93 \\
         Drivers w/ F1$>$0.9 & 34/45 & 32/45 & 3/45 & 1/45 \\
         Drivers w/ F1$<$0.5 & 1/45 & 1/45 & 19/45 & 35/45 \\
    \bottomrule
    \end{tabular}
    }
    \label{tab:driver_metrics}
\end{table}

%Figure \ref{fig:roc} presents the Receiver Operating Characteristic (ROC) curves. TEMPEST achieves a macro AUC of 99.52\%, Sup-TL 95.25\%, and DriverRep 87.56\%. The separation between curves reinforces that the ArcFace angular margin not only improves top-1 identification but also strengthens the overall decision boundary across all 45 driver classes.

% \begin{figure}[h]
%     \centering
%     \includegraphics[width=0.8\linewidth]{figs/roc_raw.png}
%     \caption{Macro-averaged ROC curves for the three deep learning models on the 45-driver DriverID dataset under temporal evaluation. TEMPEST achieves a macro AUC of 0.9952.}
%     \label{fig:roc}
% \end{figure}

\subsubsection{Cross-Dataset Evaluation on KIA}
\begin{figure}[h]
    \centering
    \includegraphics[width=0.99\linewidth]{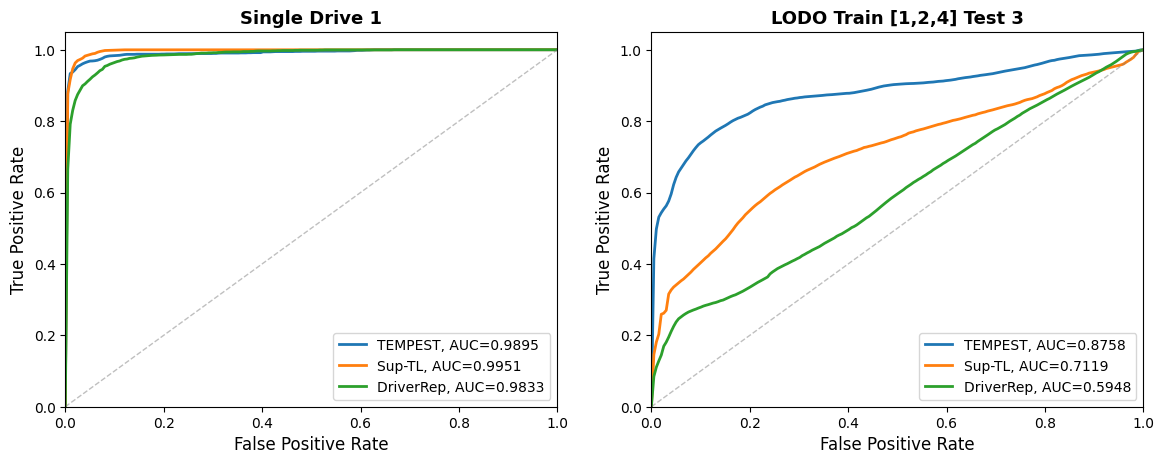}
    \caption{ROC curves for the three deep learning models on the KIA Soul dataset. Left: within-session evaluation on Drive 1. Right: cross-session LODO evaluation (train on Drives 1,2,4, test on Drive 3).}
    \label{fig:kia_roc}
\end{figure}

To assess whether TEMPEST's advantages generalize beyond the DriverID dataset, we evaluated all three models on the 10-driver HCRL KIA Soul dataset under within-session and cross-session LODO protocols. Figure~\ref{fig:kia_roc} presents ROC curves for two representative scenarios: within-session evaluation on Drive 1 and LODO evaluation with Drive 3 held out.

Within-session, all three models achieve high AUC values (TEMPEST 0.9895, Sup-TL 0.9951, DriverRep 0.9833), confirming that session-shared training and test data artificially inflates all embedding approaches equally. The cross-session protocol reveals a sharp divergence: TEMPEST maintains an AUC of 0.8758 while Sup-TL drops to 0.7119 and DriverRep collapses to 0.5948, approaching chance-level discrimination. The AUC degradation from within- to cross-session is 0.11 for TEMPEST versus 0.28 and 0.39 for Sup-TL and DriverRep, respectively, are consistent with the accuracy results in Table~\ref{tab:evaluation_impact} and confirm that ArcFace produces embeddings that retain discriminative structure across sessions, whereas triplet-based objectives overfit to session-specific patterns. The per-fold variance further underscores the challenge: TEMPEST ranges from 40.93\% (Drive 2 holdout) to 65.96\% (Drive 3 holdout), highlighting that individual sessions introduce distributional shifts that no current model fully resolves. This result motivates the need for multi-session, multi-route dataset collection as a priority for the field.

\subsection{Limitations and Practical Considerations}

The primary limitations of this work are dataset scope and evaluation coverage. DriverID provides a single session per driver on a fixed route, preventing direct cross-session generalization analysis on this cohort, while the KIA dataset, though offering four sessions per driver, covers only 10 drivers; together, the two datasets provide complementary but incomplete coverage of the real-world deployment space. Beyond data, physiological sensing is limited to a single wrist-worn heart rate monitor, and richer multi-modal signals may further improve inter-driver discrimination. Finally, all experiments assume a closed-set enrollment, leaving open-set rejection of unknown drivers as an important open problem for practical deployment.

Despite these limitations, TEMPEST is well-suited for real-world deployment: 720K parameters, a 2.80~MB footprint, and 50-epoch convergence (22 minutes on a single NVIDIA GB10 GPU) make it feasible for resource-constrained vehicular environments. The decoupled encoder--embedding inference architecture eliminates retraining entirely.

Finally, while driving-behavior embeddings enable continuous, passive verification without physical biometrics, they raise privacy considerations. All data used in this work was collected under IRB approval with explicit participant consent. At inference time, only the 96-dimensional embedding vector is stored; no raw sensor signals are retained, substantially limiting sensitive data exposure.

\section{Conclusion}

This work introduces TEMPEST, a TCN-based embedding model trained with an additive angular margin (ArcFace) loss for scalable driver identification. Because TEMPEST and DriverRep share the same encoder architecture and inference pipeline, their comparison constitutes a direct ablation of training objective: replacing triplet loss with ArcFace yields a 58.4~pp Rank-1 accuracy gain (33.35\% $\rightarrow$ 91.71\%) under identical temporal evaluation conditions, establishing the training objective as the dominant factor in behavioral biometric embedding quality. TEMPEST's cosine-similarity inference scheme supports truly open enrollment, as new drivers are added by updating the embedding gallery with no retraining or classifier refitting of any kind. Across both the 45-driver DriverID and 10-driver KIA Soul datasets, TEMPEST degrades gracefully with pool size and retains discriminative structure across sessions, whereas triplet-based alternatives collapse under the same conditions. Critically, we demonstrate that random cross-validation inflates accuracy by up to 80~pp relative to chronological and cross-session evaluations, rendering prior comparisons unreliable. The results reported here establish realistic, reproducible baselines for the community under rigorous, leakage-aware protocols. Future work should prioritize multi-session, multi-route data collection to strengthen cross-session generalization; explore adaptive angular margins for commonly confused driver pairs; and investigate multi-modal attention mechanisms that optimally fuse vehicle dynamics and physiological signals.

%%%%%%%%%%%%%%%%%%%%%%%%%%%%%%%%%%%%%%%%%%%%%%%%%%%%%%%%%%%%%%%

\section*{Acknowledgment}
This manuscript has been authored in part by UT-Battelle, LLC, under contract DE-AC05-00OR22725 with the US Department of Energy (DOE). The publisher acknowledges the US government license to provide public access under the DOE Public Access Plan (https://energy.gov/doe-public-access-plan). This work is in support of our Domestic Transportation Security Program at ORNL. 
This work was supported in part by institutional research funding and collaborative contributions from participating drivers and data collection teams. The authors thank the open-source community for providing benchmark datasets and tooling that enabled reproducible evaluation.

\section*{Dataset and Code Dissemination}
The raw DriverID dataset is publicly accessible upon request through the Oak Ridge National Laboratory data repository~\cite{driverdata}. To support full reproducibility, all code accompanying this paper, including the TEMPEST model, training pipeline, preprocessing scripts to generate the cleaned 45-driver dataset, and evaluation protocols, will be released on a public GitHub repository upon acceptance.

% %------------------------------------------------------------------------
% \section{Only for Final copy}

% You must include your signed IEEE copyright release form when you submit
% your finished paper. We MUST have this form before your paper can be
% published in the proceedings.

{\small
\bibliographystyle{ieee}
\bibliography{references}

@inproceedings{Enev2016,
  author    = {M. Enev and A. Takakuwa and K. Koscher and T. Kohno},
  title     = {Automobile driver fingerprinting},
  booktitle = {Proc. Priv. Enhancing Technol.},
  volume    = {2016},
  number    = {1},
  pages     = {34--50},
  year      = {2016},
  note = {\url{https://doi.org/10.1515/popets-2015-0029}}
}

@article{Miyajima2007,
  author  = {C. Miyajima and Y. Nishiwaki and K. Ozawa and T. Wakita and K. Itou and K. Takeda and F. Itakura},
  title   = {Driver modeling based on driving behavior and its evaluation in driver identification},
  journal = {Proc. IEEE},
  volume  = {95},
  number  = {2},
  pages   = {427--437},
  year    = {2007},
  note = {\url{https://ui.adsabs.harvard.edu/link_gateway/2007IEEEP..95..427M/doi:10.1109/JPROC.2006.888405}}
}

@inproceedings{Choi07,
  author    = {S. Choi and J. Kim and D. Kwak and P. Angkititrakul and J. H. L. Hansen},
  title     = {Analysis and classification of driver behavior using in-vehicle {CAN}-bus information},
  booktitle = {Proc. Bienn. Workshop DSP In-Vehicle Mobile Syst.},
  year      = {2007},
  note = {\url{https://www.researchgate.net/publication/228936619_Analysis_and_Classification_of_Driver_Behavior_Using_in-Vehicle_CAN-BUS_Information}}
}

@inproceedings{Wakita06,
  author    = {T. Wakita and K. Ozawa and C. Miyajima and K. Igarashi and K. Itou and K. Takeda and F. Itakura},
  title     = {Driver identification using driving behavior signals},
  booktitle = {Proc. IEEE Intell. Transp. Syst. Conf.},
  pages     = {396--401},
  year      = {2006},
  note = {\url{https://doi.org/10.1093/ietisy/e89-d.3.1188}}
}

@article{Wahab09,
  author    = {A. Wahab and C. Quek and C. K. Tan and K. Takeda},
  title     = {Driving profile modeling and recognition based on soft computing approach},
  journal   = {IEEE Trans. Neural Netw.},
  volume    = {20},
  number    = {4},
  pages     = {563--582},
  year      = {2009},
  note = {\url{https://doi.org/10.1109/tnn.2008.2007906}}
}

@article{Ezzini18,
  author  = {S. Ezzini and I. Berrada and M. Ghogho},
  title   = {Who is behind the wheel? {D}river identification and fingerprinting},
  journal = {J. Big Data},
  volume  = {5},
  number  = {1},
  pages   = {9},
  year    = {2018},
  note = {\url{https://doi.org/10.1186/s40537-018-0118-7}}
}

@article{Mekki,
  author  = {Abdellah {El Mekki} and Afaf Bouhoute and Ismail Berrada},
  title   = {Improving Driver Identification for the Next-Generation of In-Vehicle Software Systems},
  journal = {IEEE Transactions on Vehicular Technology},
  volume  = {68},
  number  = {8},
  pages   = {7406--7415},
  year    = {2019},
  note = {\url{https://doi.org/10.1109/TVT.2019.2924906}}
}

@article{Xun_2020_TII,
  author  = {Y. Xun and J. Liu and N. Kato and Y. Fang and Y. Zhang},
  title   = {Automobile driver fingerprinting: A new machine learning approach},
  journal = {IEEE Transactions on Industrial Informatics},
  volume  = {16},
  number  = {2},
  pages   = {1417--1426},
  year    = {2020},
  note = {\url{https://doi.org/10.1109/TII.2019.2946626}}
}

@article{Zhang_2019_Sensors,
  author  = {J. Zhang and Z. Wu and F. Li and C. Xie and T. Ren and J. Chen and L. Liu},
  title   = {A deep learning framework for driving behavior identification on in-vehicle CAN-bus sensor data},
  journal = {Sensors},
  volume  = {19},
  number  = {6},
  pages   = {1356},
  year    = {2019},
  note = {\url{https://doi.org/10.3390/s19061356}}
}

@article{Zhengping,
  author  = {Li, Zhengping and Zhang, Kai and Chen, Bokui and Dong, Yuhan and Zhang, Lin},
  title   = {Driver identification in intelligent vehicle systems using machine learning algorithms},
  journal = {IET Intelligent Transport Systems},
  volume  = {13},
  pages   = {40-47},
  year    = {2019},
  note = {\url{https://doi.org/10.1049/iet-its.2017.0254}}
}

@inproceedings{girma2019driveridentificationbasedvehicle,
  title={Driver identification based on vehicle telematics data using LSTM-recurrent neural network},
  author={Girma, Abenezer and Yan, Xuyang and Homaifar, Abdollah},
  booktitle={2019 IEEE 31st International Conference on Tools with Artificial Intelligence (ICTAI)},
  pages={894--902},
  year={2019},
  organization={IEEE}
}

@article{ABDENNOUR2021103707,
  author  = {N. Abdennour and T. Ouni and N. B. Amor},
  title   = {Driver identification using only the CAN-bus vehicle data through an RCN deep learning approach},
  journal = {Robot. Auton. Syst.},
  volume  = {136},
  pages   = {103707},
  year    = {2021},
  note = {\url{https://doi.org/10.1016/j.robot.2020.103707}}
}

@article{Alqahtani,
  author  = {Alqahtani, Hamed and Kumar, Gulshan},
  title   = {A deep learning-based intrusion detection system for in-vehicle networks},
  journal = {ACM Comput. Electr. Eng.},
  volume  = {104},
  number  = {PB},
  numpages = {16},
  year    = {2022},
  note = {\url{https://doi.org/10.1016/j.compeleceng.2022.108447}}
}

@inproceedings{Hallac18,
  title={Drive2vec: Multiscale state-space embedding of vehicular sensor data},
  author={Hallac, David and Bhooshan, Suvrat and Chen, Michael and Abida, Kacem and Leskovec, Jure and others},
  booktitle={2018 21st International Conference on Intelligent Transportation Systems (ITSC)},
  pages={3233--3238},
  year={2018},
  organization={IEEE}
}

@article{driver2vec,
  title={Driver2vec: Driver identification from automotive data. arXiv 2021},
  author={Yang, J and Zhao, R and Zhu, M and Hallac, D and Sodnik, J and Leskovec, J},
  journal={arXiv preprint arXiv:2102.05234},
  year={2021}
}

@inproceedings{Hallac16,
  author    = {D. Hallac and S. Sharang and R. Stahlmann and A. Lamprecht and M. Huber and M. Roehder and J. Leskovec},
  title     = {Driver identification using automobile sensor data from a single turn},
  booktitle = {2016 IEEE 19th International Conference on Intelligent Transportation Systems (ITSC)},
  pages     = {953--958},
  year      = {2016},
  note = {\url{https://doi.org/10.1109/ITSC.2016.7795670}}
}

@inproceedings{DriverRep2020,
  author    = {M. Nasr Azadani and A. Boukerche},
  title     = {{DriverRep}: Driver identification through driving behavior embeddings},
  booktitle = {Journal of Parallel and Distributed Computing},
  volume = {162},
  pages = {105--117},
  year      = {2020},
  note = {\url{https://doi.org/10.1016/j.jpdc.2022.01.010}}
}

@article{MARTINELLI2020102504,
  author  = {Fabio Martinelli and Francesco Mercaldo and Albina Orlando and Vittoria Nardone and Antonella Santone and Arun Kumar Sangaiah},
  title   = {Human behavior characterization for driving style recognition in vehicle system},
  journal = {Computers and Electrical Engineering},
  volume  = {83},
  pages   = {102504},
  year    = {2020},
  note = {\url{https://doi.org/10.1016/j.compeleceng.2017.12.050}}
}

@misc{driverdata,
  author = {{Oak Ridge National Laboratory}},
  title  = {{ORNL} Driver Identification Dataset},
  year   = {2023},
  note   = {Available upon request},
  howpublished = {\url{https://doi.org/10.13139/ORNLNCCS/2513388}}
}

@misc{kia2016,
  author    = {Kwak, Byung Il and Woo, Jiyoung and Kim, Huy Kang},
  title     = {Know Your Master: Driver Profiling-based Anti-theft Method},
  year      = {2016},
  howpublished = {\url{https://sites.google.com/hksecurity.net/hcrl/Datasets/driving-dataset}}
}

@misc{KIASoul2017,
  author = {{HCRL}},
  title  = {{KIA Soul} driving dataset},
  year   = {2017},
  note   = {Publicly available at \url{https://ocslab.hksecurity.net/}},
}

@misc{canbus,
  author = {{Robert Bosch GmbH}},
  title  = {{CAN} specification version 2.0},
  year   = {1991},
}

@misc{vbox,
  author = {{Racelogic}},
  title  = {{VBOX} 3i: 100 {Hz} {GPS} data logger},
  year   = {2020},
  note   = {\url{https://www.vboxautomotive.co.uk/}}
}

@misc{hr,
  author = {{Empatica}},
  title  = {{E4} wristband: Real-time physiological data streaming and visualization},
  year   = {2020},
  note   = {\url{https://www.empatica.com/research/e4/}}
}

@article{wevj13110207,
  author  = {Dengfeng Zhao and Junjian Hou and Yudong Zhong and Wenbin He and Zhijun Fu and Fang Zhou},
  title   = {Driver Identification Methods in Electric Vehicles, a Review},
  journal = {World Electr. Veh. J.},
  volume  = {13},
  number  = {11},
  pages   = {207},
  year    = {2022},
  note = {\url{https://doi.org/10.3390/wevj13110207}}
}

@article{khan2022machinelearningapproachdriver,
  title={A machine learning approach for driver identification based on CAN-BUS sensor data},
  author={Khan, Md Abbas Ali and Ali, Mphammad Hanif and Haque, AKM and Habib, Md Tarek},
  journal={arXiv preprint arXiv:2207.10807},
  year={2022}
}

@article{Khan2022,
  author  = {Muhammad Qasim Khan and Sukhan Lee},
  title   = {A comprehensive survey of driving monitoring and assistance systems},
  journal = {Sensors},
  volume  = {19},
  number  = {11},
  year    = {2019},
  note = {\url{https://doi.org/10.3390/s19112574}}
}

@article{Zeng22,
  author  = {Li Zeng and Mohamad al-Rifai and Sergiu Chelaru and Michael Nolting and Wolfgang Nejdl},
  title   = {On the Importance of Contextual Information for Building Reliable Automated Driver Identification Systems},
  journal = {Proc. 2020 IEEE 23rd International Conference on Intelligent Transportation Systems (ITSC)},
  pages   = {1--6},
  year    = {2020},
  note = {\url{https://doi.org/10.1109/ITSC45102.2020.9294439}}
}

@article{TCN2018,
  author  = {S. Bai and J. Z. Kolter and V. Koltun},
  title   = {An empirical evaluation of generic convolutional and recurrent networks for sequence modeling},
  journal = {arXiv preprint arXiv:1803.01271},
  year    = {2018},
  note = {\url{https://doi.org/10.48550/arXiv.1803.01271}}
}

@article{wavenet,
  author  = {A. {van den Oord} and S. Dieleman and H. Zen and K. Simonyan and O. Vinyals and A. Graves and N. Kalchbrenner and A. Senior and K. Kavukcuoglu},
  title   = {{WaveNet}: A generative model for raw audio},
  journal = {arXiv preprint arXiv:1609.03499},
  year    = {2016},
  note = {\url{https://doi.org/10.48550/arXiv.1609.03499}}
}

@inproceedings{ArcFace2019,
  author    = {Deng, Jiankang and Guo, Jia and Yang, Jing and Xue, Niannan and Kotsia, Irene and Zafeiriou, Stefanos},
  title     = {{ArcFace}: Additive angular margin loss for deep face recognition},
  booktitle = {IEEE Transactions on Pattern Analysis and Machine Intelligence},
  year={2022},
  volume={44},
  number={10},
  pages={5962-5979},
  note = {\url{https://doi.org/10.1109/TPAMI.2021.3087709}}
}

@inproceedings{FaceNet,
  author    = {Schroff, Florian and Kalenichenko, Dmitry and Philbin, James},
  title     = {{FaceNet}: A unified embedding for face recognition and clustering},
  booktitle = {2015 IEEE Conference on Computer Vision and Pattern Recognition (CVPR)},
  pages     = {815--823},
  year      = {2015},
  note = {\url{https://doi.org/10.1109/CVPR.2015.7298682}}
}

@article{tSNE2008,
  author  = {L. van der Maaten and G. Hinton},
  title   = {Visualizing data using {t-SNE}},
  journal = {J. Mach. Learn. Res.},
  volume  = {9},
  pages   = {2579--2605},
  year    = {2008},
  note = {\url{http://www.jmlr.org/papers/v9/vandermaaten08a.html}}
}

@inproceedings{Biggs2024,
  author    = {T. Biggs and T. Lanigan and D. Ruddell and E. E. Gallegos and J. Daily},
  title     = {Modeling a Heavy-duty Vehicle Data Collection Process for Authenticating Driver Identity and Analyzing Driver Behavior Under Duress},
  booktitle = {Proc. 19th Annu. System of Systems Eng. Conf. (SoSE)},
  pages     = {256--263},
  year      = {2024},
  note = {\url{https://doi.org/10.1109/SOSE62659.2024.10620958}}
}
}

\end{document}